\pdfoutput=1

\documentclass[11pt]{article}
\usepackage{emnlp2021}

\usepackage{times}
\usepackage{latexsym}
\usepackage{tabularx}
\usepackage{booktabs}
\usepackage{float}
\usepackage{graphicx}
\usepackage{hyperref}
\usepackage{url}
\usepackage{microtype}

\usepackage[T1]{fontenc}

\usepackage[utf8]{inputenc}

\usepackage{microtype}
\newcommand\blfootnote[1]{%
  \begingroup
  \renewcommand\thefootnote{}\footnote{#1}%
  \addtocounter{footnote}{-1}%
  \endgroup
}

\title{``So You Think You're Funny?'': Rating the Humour Quotient in Standup Comedy}

\author{
Anirudh Mittal$^{\dagger}$,
Pranav Jeevan$^{\diamond}$, 
Prerak Gandhi$^{\clubsuit}$, 
Diptesh Kanojia$^{\ddagger}$, 
Pushpak Bhattacharyya$^{\star}$ \\
  $^{\dagger,\diamond,\clubsuit,\star}$Indian Institute of Technology Bombay, Mumbai \\
  $^{\ddagger}$Centre for Translation Studies, University of Surrey, United Kingdom\\
  $^{\dagger,\clubsuit,\star}$\texttt{\{anirudhmittal,prerakgandhi,pb\}@cse.iitb.ac.in}\\
  $^{\diamond}$\texttt{pranavjp@ee.iitb.ac.in}\\
  $^{\ddagger}$\texttt{d.kanojia@surrey.ac.uk}}

\begin{document}
\maketitle
\begin{abstract}
Computational Humour (CH) has attracted the interest of Natural Language Processing and Computational Linguistics communities. Creating datasets for automatic measurement of humour quotient is difficult due to multiple possible interpretations of the content. In this work, we create a multi-modal humour-annotated dataset ($\sim$40 hours) using stand-up comedy clips. We devise a novel scoring mechanism to annotate the training data with a humour quotient score using the audience's laughter. The normalized duration (laughter duration divided by the clip duration) of laughter in each clip is used to compute this humour coefficient score on a five-point scale (0-4). This method of scoring is validated by comparing with manually annotated scores, wherein a quadratic weighted kappa of 0.6 is obtained. We use this dataset to train a model that provides a ``funniness'' score, on a five-point scale, given the audio and its corresponding text. We compare various neural language models for the task of humour-rating and achieve an accuracy of $0.813$ in terms of Quadratic Weighted Kappa (QWK). Our ``Open Mic'' dataset is released for further research along with the code.
\end{abstract}

\section{Introduction}

\blfootnote{$^{\dagger}$Corresponding Author}Humour is one of the most important lubricants of communication between people. Humour is subjective and, at times, also requires cultural knowledge as humour is often dependent on stereotypes in a culture or a country. At times, even cultural appropriation is used to convey humour, which can be offensive to minority cultures\footnote{\href{https://www.vox.com/the-highlight/2020/1/15/21065939/comedy-racism-asian-american-rosie-odonnell-shane-gillis-awkwafina-ali-wong}{Racism in Comedy: An opinion piece.}}~\citep{book,book1}. 
The factors listed above, along with the underlying subjectivity in humour render the task of rating humour, difficult for machines~\citep{meaney-2020-crossing}. The task of humour classification suffers due to this subjectivity and the lack of datasets that rate the ``funniness'' of content.

In this paper, we propose rating humour on a scale of zero to four. We create the first multi-modal dataset\footnote{\href{https://github.com/TheExtraSemiColon/AI-OpenMic}{Dataset and Code}} using standup comedy clips and compute the humour quotient of each clip using the audience laughter. The validity of our scoring criteria is verified by finding the overall agreement between human annotation and automated scores. We use the audio and text-based signals to process this multi-modal data to generate `humour ratings'. Since humour annotation is subjective, even the data annotated by humans might not provide an objective measure. We reduce this subjectivity by taking laughter feedback from a larger audience. To the best of our knowledge, no previous literature has proposed an automatically humour-rated multi-modal dataset and used it in ML model-building to automatically obtain the humor score.

Standup comedy is an art form where the delivery of humour has a much larger context, and there are multiple jokes and multiple related punchlines in the same story. The resulting laughter from the audience depends on various factors, including the understanding of the context, delivery, and tonality of the comic. Standup comedy seems to be an ideal choice for a humour rating dataset as it inherently contains some feedback in terms of the audience laughter. We believe a smaller context window restricts computational models, but we know this is not the case for the human audience. Hence, our approach \textit{utilises live audience laughter as a measure to rate the humour quotient} in the data created. We also believe that such an approach can generate insights into what aspects of stories and their delivery make them funny.

Our humour rating model is partly inspired by the character ``TARS'' from the movie ``Interstellar'', which generates funny responses based on adjustable humour setting~\citep{film}. An essential step in developing such a machine that can adjust its ``funniness'' is to create a model that can recognize and rate the ``funniness'' of a joke. With this work, we aim to release a dataset that can help researchers shed light on the humour quotient of a particular text. \textbf{The key contributions of this paper are:} (a) Creation and public release of an automatically rated multi-modal dataset based on English standup comedy clips and (b) Manual evaluation of this dataset along with humour-rating quotient defined on a Likert-scale~\citep{likert1932technique}. 

\section{Related Work}

Most of the previous work on computational humour has been towards the detection of humour. Smaller joke formats like one-liners which have just a single line of context, have been used~\citep{journal1}.  Language models like BERT are used for generating sentence embeddings, which have been shown to outperform other architectures in humour detection on short texts~\citep{annamoradnejad2020colbert}. Since humour depends on how the speaker's voice changes, the audio features, and language features have been used as inputs for machine learning models for humour detection.~\citet{bertero-fung-2016-deep} use audio and language features to detect humour in The Big Bang Theory sitcom dialogues.~\citet{Park2018LaughbotDH} passed audio and language features from a conversation dataset into an RNN to create a chatbot that can detect and respond to humour.~\citet{hasan-etal-2019-ur} built a multi-modal dataset that uses text, audio, and video inputs for humour detection. There are existing datasets that rate the humour in tweets and Reddit posts, with the help of human annotators~\citep{miller2020ofaiukp,castro-etal-2018-crowd,weller2019humor}. Creating human-annotated datasets is costly in terms of both time and money and has been one of the noted issues for creating humour datasets.~\citet{Yang20192,Yang2019} used time-aligned user comments for generating automated humour labels for multi-modal humour identification tasks and found good agreement with manually annotated data. However, none of the previously existing datasets are created with standup comedy clips.

We present the first multi-modal dataset that uses a non-binary rating system. We use standup comedy clips which makes our dataset scalable and diverse. The dataset is novel in terms of the use of long contextual jokes ($\sim 2$ mins) and audience laughter which helps annotate the funniness in each clip in an automated manner.

\section{Dataset Acquisition and Pre-processing}
\label{sec:length}

In this section, we describe the creation of our multi-modal dataset and the manual evaluation performed with the help of human annotators.

We gather 36 English language standup comedy shows from 32 comedians available on the web, where the length of each original clip is $\sim 1$ hour. We further segment them manually into $927 \sim 2$ minute long clips. The standups are chosen based on the clarity of the audience feedback laughter. We choose comics from diverse categories of gender, nationality, and culture to ensure representation and reduce bias. While segmenting them, we ignore the clips, which results in laughter on interaction with the audience/personal jokes. We also create text files with the transcript for each audio clip from multiple online sources~\citep{Transcript}. We collect data for ``unfunny'' samples by gathering TED talk audio clips with similar speech delivery modes like standup comedy. We also segment them into 128 $\sim 2$ minute audio clips and create text files of their transcript\footnote{\href{https://www.ted.com/}{TED Talks}}. 

Clips were manually trimmed from the complete audio such that the entire context for the joke is available within the clip. This results in the overall set of $\sim2$ minute clips described above. Finally, we acquire 519 $\sim2$ minute audio clips and corresponding transcript text files in our dataset. The train-test split is chosen to be 70-30.

\subsection{Laughter Detection}
\label{subsec:ld}
To find the humour quotient rating of each clip, we use the feedback of the audience laughter as discussed above. We measure the intensity and recorded time intervals of audience laughter in the clip~\citep{detection}. We modify this library to output \textit{the sum of the duration, of all laughs in the clip}. Based on hyperparameter tuning, we set the threshold parameters, adjusting the minimum probability for laughter detection to 0.7. Further, the minimum laughter duration parameter is set to 0.1. This allows us to get the humour quotient from the total duration of the audience laughter in the clip. 

\subsection{Scoring Humour Quotient}
The sum of the duration of all the laugh intervals is detected from each clip. Since longer clips tend to have more jokes and hence a higher score, we eliminate this bias by dividing the sum with the duration of the clip. We use a Likert-scale to regard for the subjectivity in human opinion on each clip. The mean $\mu$ and standard deviation $\sigma$ of all the scores are calculated. A rule for assigning a 5 point rating (0-4) for each clip is devised as shown in Table~\ref{tab:scoring} (Column 3).  The number of samples for each class in our rating system is shown in Table~\ref{tab:scoring}. 

\begin{table}[t!]
\centering
\resizebox{0.8\columnwidth}{!}{%
\begin{tabular}{@{}ccl@{}}
\toprule
\textbf{Rating} & \textbf{\# Clips} & \textbf{Scoring Criteria} \\ \midrule
4 & 233 & score $>$ $\mu$ + 0.75$\sigma$ \\
3 & 185 & $\mu$ + 0.75$\sigma$ $\geq$ score $>$ $\mu$ \\
2 & 256 & $\mu$ $\geq$ score $>$ $\mu$ - 0.75$\sigma$ \\
1 & 253 & $\mu$ - 0.75$\sigma$ $\geq$ score $>$ 0  \\
0 & 128 & score $=$ 0 \\ \bottomrule
\end{tabular}%
}
\\
\caption{Number of clips and the scoring criteria for assigning humour rating to each clip based on the mean ($\mu$) and standard deviation ($\sigma$) of the scores}
\label{tab:scoring}
\end{table}

\subsection{Human Annotation}
Three human annotators (2 males, 1 female) between the ages of 21-33 are assigned to rate the humour quotient in our dataset. The annotators are instructed to rate each clip based \textit{solely on the audience laughter feedback} rather than their perception of the humour quotient of the clip. This allows the annotators to be \textit{unbiased towards a particular comedian or humour genre}. The annotations were performed in a \textit{closed-room environment, without any external noise}. 

\section{Experiment Setup and Methodology}

In this section, we describe the features used for the humour rating prediction task along with the additional pre-processing in detail.

\subsection{Network Architecture}
\begin{figure}[ht!]
\centering
\includegraphics[width=0.99\columnwidth]{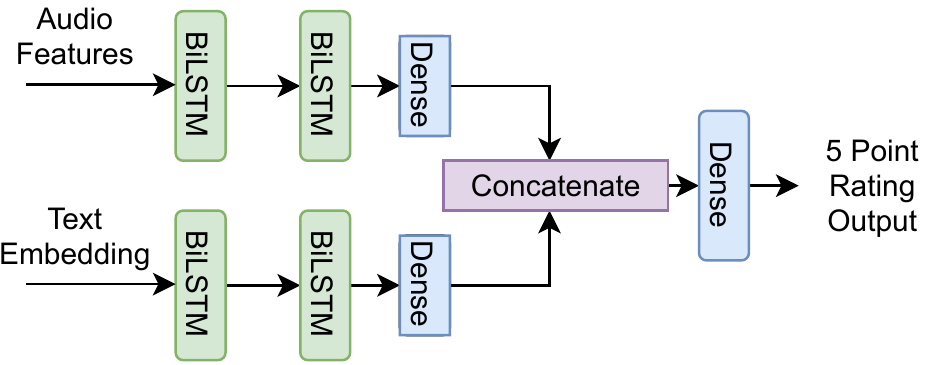}
\caption{Neural Network Architecture}
\label{fig:NN}
\end{figure}

The text embeddings and audio features are given as input to separate Bi-LSTM layers followed by separate, Dense layers~\citep{bilstm} as shown in Figure~\ref{fig:NN}. The output from these two pathways is then concatenated and fed to a classifier that outputs one-hot encoding of the 5-point rating.

\subsection{Muting Laughter}

Before extracting audio features, we remove the audience laughter and isolate the speaker's voice from each clip. Retaining the audience laughter may enable a neural network to utilize it and predict a score without using information from the text and other audio features. We envision creating a system that can predict the funniness of any clip. Such clips will not have laughter as an indicator, so we train and test our model on the muted clips. Please note that laughter is extracted separately to generate the funniness score (Section \ref{subsec:ld}). We use~\citet{mute} to mute audience laughter from audio segments, thus, resulting in clips that are then used for extracting audio features.

\subsection{Audio Features}
Audio features such as MFCCs, RMS energy, and Spectrogram are extracted from the laughter-muted clips~\citep{brian_mcfee_2020_3955228}. These 3 feature tensors are concatenated to create a single feature of dimension 33 for each time sample. The maximum sequence length for audio embeddings was set as 8000. The clips with a lesser duration were padded with zeroes for uniformity. These features convey information about the volume, intonation, and emotion of the speaker, which are important for humour.  

\subsection{Textual Features}
Additionally, we use the textual features extracted from various language models to ensure that the context of each joke is retained. We use BERT-derived models to generate contextual embeddings for each clip, which ensure attention over the entire text of the clip~\citep{wolf-etal-2020-transformers}. BERT-derived models can process sequences of token length 512; thus, we employ them for the entire transcript of each $\sim 2$ minute clip. We sum the output of the final 4 layers from these models to obtain a clip embedding~\citep{bertembedding}.

As baseline textual features, we use GloVe embeddings~\citep{inproceedings}. For obtaining textual features, we experiment with BERT$_{base}$, BERT$_{large}$, XLM, DistilBERT, RoBERTa$_{base}$ and RoBERTa$_{large}$ to generate text embeddings~\citep{DBLP:journals/corr/abs-1810-04805,lample2019cross,Sanh2019DistilBERTAD,DBLP:journals/corr/abs-1907-11692}. 

\subsection{Methodology}

The audio features and textual features are fed as input to the network for obtaining an output rating on the scale of 0-4. To evaluate our approach for scoring each clip, we obtain Cohen's weighted Kappa with quadratic weights, \textit{i.e.,}  Quadratic Weighted Kappa score (QWK)~\citep{cohen1968weighted}   between our scoring mechanism (Table~\ref{tab:scoring}) and the model output. We use QWK as a scoring mechanism because, unlike accuracy and F-Score, it considers that the system may randomly assign a particular label to a clip. The QWK score also penalizes mismatches more than linear or unweighted Kappa by taking the quadratic weights into account. Additionally, we validate the scores provided by our scoring mechanism by obtaining QWK with the human annotation performed. 
\begin{table}[ht!]
\centering
\resizebox{0.98\columnwidth}{!}{%
\begin{tabular}{@{}lc@{}}
\toprule
\multicolumn{2}{c}{\textbf{Pairwise Agreement}} \\ \midrule
Annotators A and B & $0.643$ \\
Annotators B and C & $0.926$ \\
Annotators C and A & $0.611$ \\
Average pairwise Cohen's Kappa & $0.634$ \\ \midrule
Fleiss' Kappa & $0.632$ \\
\textbf{Krippendorff's alpha} & \textbf{$0.632$}\\\bottomrule
\end{tabular}%
}
\\
\caption{Inter-Annotator Agreement (Fleiss' Kappa and Krippendorff's alpha) values along with pairwise agreement among the annotators}
\label{tab:annotator}
\end{table}

\begin{table}[t!]
\centering
\resizebox{0.65\columnwidth}{!}{%
\begin{tabular}{ll}
\hline
\textbf{Annotaters} & \textbf{QWK} \\ \hline
Human A & $0.659$ \\
Human B & $0.562$ \\
Human C & $0.563$ \\
\textbf{Average} & \textbf{$0.595$} \\ \hline
\textbf{Textual Features} & \textbf{QWK} \\ \hline
GloVe & $0.691$ \\
BERT$_{base}$ & $0.722$ \\
BERT$_{large}$ & $0.796$ \\
DistilBERT & $0.721$ \\
RoBERTa$_{base}$ & $0.775$ \\
\textbf{RoBERTa$_{large}$} & \textbf{$0.813$} \\
XLM & $0.714$ \\ \hline
\end{tabular}%
}
\\
\caption{(a) Quadratic Weighted Kappa (QWK) scores between the scores provided by human annotators, and our scoring mechanism (b) QWK scores between the various language models combined with neural networks, and our scoring mechanism.}
\label{tab:results}
\end{table}
\vspace{-0.1cm}
\section{Results}

In Table~\ref{tab:annotator}, we show the Krippendorff alpha, Fleiss' Kappa, and pairwise agreement between human annotators~\citep{krippendorff04,cohen1960,fleiss1971mns}. The inter-annotator agreement between any two annotators is above 0.60, which signifies ``substantial'' agreement between them~\citep{articleFleiss}. We evaluate our scoring mechanism by comparing it with the manually annotated data by human annotators, as shown in Table~\ref{tab:results}. An average QWK of 0.595 was observed, indicating significant agreement with them~\citep{Kappa}. 

Table~\ref{tab:results} also shows the QWK among the neural network outputs with our scoring mechanism. With the neural network output, we see a significant agreement when compared with our scoring mechanism. Even the GloVe-based model performs reasonably well when matched with our scoring mechanism. Embeddings created from BERT-derived language models showed considerable improvement from baseline performance. RoBERTa$_{large}$ outperforms all the other language models and shows an improvement of 12\% points over the baseline GloVe score. Since RoBERTa is pre-trained on datasets that contain text in a story-like format similar to standup comedy text~\citep{DBLP:journals/corr/abs-1907-11692}, RoBERTa$_{large}$ can be seen performing better than all the other textual features. Analysing the confusion matrix of these models shows that RoBERTa$_{large}$ and BERT$_{large}$ can distinguish different levels of humourousness quite well. They show the highest accuracy in identifying the non-funny clips. DistilBERT could not perform as well as BERT$_{large}$ because humour needs better quality text embeddings to understand the full context, which DistilBERT cannot provide due to the lower number of parameters in the model. 

Larger models with embedding dimensions of 1024 (BERT$_{large}$, RoBERTa$_{large}$) and 2048 (XLM) performed better than smaller models. A larger neural network would need a dataset of significant size to train, which also shows that our dataset is reasonably sized. For BERT$_{base}$, when we increased the Bi-LSTMs in the initial layer from 256 to 512, we see a slight improvement in the Quadratic Weighted Kappa value which shows that larger embeddings need a bigger neural network to classify accurately.

We further probe our best-performing model with an ablation test and observe that audio-based features (0.66 QWK) outperform text-based features (0.48 QWK). This contradicts what was observed by~\citet{hasan-etal-2019-ur} as humour in standup often depends on the tonality and the well-enunciated punchlines.

\section{Discussion}

Analysis of the predicted ratings show that our model can identify non-funny clips and most funny clips with very high accuracy. In cases of error in assigning ratings to the intermediate funny clips, the assigned ratings are not off by more than one rating point, for \textit{e.g.,} a clip rated 3 is assigned a rating of 2 or 4. This error should not be considered as a failure of our model since assigning a precise funniness rating in a definite way to intermediate funny clips is hard even for humans. So it is reasonable to expect our model to commit similar errors in assigning the ratings as a human would. In the individual confusion matrices obtained for both feature sets, we observe the maximum incorrect predictions among the classes 2/3 and 3/4. We correlate these results with the human annotation and observe that even human annotators differ mostly in these two classes. All our annotators observed that clips with a moderate amount of laughter could be rated either as 2 or as 3, since such annotations are difficult to be discerned to a particular class. 

Additionally, we observe that there are only $16/1055$ cases where none of the ratings of the three human annotators match with each other. Out of which, only one rating differs (4, 1, 2) with a difference of $>=$ 2. In the other 15 ratings, the difference between the highest and lowest human ratings is $<=$ 2 (\textit{e.g.,} 4-2-3). There are around $\sim$400 cases where 2 annotators fully agree. The rest of the $\sim$600 ratings are where all three annotators fully agree in their ratings for each clip shown to them.

As we evaluate the clips misclassified by our model, we observe that 1) sarcastic and ironic statements generate human laughter, but our model does not detect it, 2) a certain kind of jokes which are morbid also categorized as ``dark humour'' is consistently classified with lower scores, whereas there is a lot of human laughter generated during such jokes, 3) subtle comparisons, for example, the usage of internet to smoking where the comedian tries to imply that both are harmful to health; are classified as ``mildly funny (1 or 2)'' by our model. 

We further evaluate clips with human annotation score difference $>$ 2. Despite providing detailed guidelines, which required our annotators to focus only on audience laughter, they have possibly focused on the content. Due to this subjectivity, we believe that our annotators may have misclassified a few clips. We trace the reasons to 1) country-specificity, leading to limited comprehension by the annotator, or 2) insensitivity towards the feelings of females, or 3) bias against a country/race which leads to the diminished absorption of the joke. This observation validates our initial discussion on the subjectivity of humorous content, along with the observation that Annotator A (female) has consistently scored such clips lower than Annotator B / C (males). However, we also observed that the audience laughter in such clips is more consistent with the scores provided by Annotators B and C.

\section{Conclusion and Future Work}
We propose a novel scoring mechanism to show that humour rating can be automated using audience laughter, which concurs well with the humour perception of humans. We create a multi-modal (audio \& text) dataset for the task of humour rating. With the help of three human annotators, we manually evaluate our scoring mechanism and show a substantial agreement in terms of QWK. Our evaluation shows that our scoring mechanism can be emulated with the help of pre-existing language models and traditional audio features. Our neural network-based experiments show that the output obtained using various language models like RoBERTa show an agreement with our scoring mechanism. Despite the inherent subjectivity in humour and its different perceptions among humans, we propose a method to rate humour and release this dataset under the CC-BY-SA-NC 4.0 license for further research. 

In the future, we would like to evaluate this scoring mechanism with the help of more human annotators. We aim to extend the dataset with the help of more standup comedy clips. Further experiments can be conducted to compare the contribution of audio, video and text features with a more detailed analysis. We would also like to perform experiments by including more audio features like Line Spectral Frequencies, Zero-Crossing rate, and Delta Coefficients. With the release of this dataset, we hope that research in computational humour can be taken further.

\bibliography{anthology,emnlp2021}

\begin{thebibliography}{33}
\expandafter\ifx\csname natexlab\endcsname\relax\def\natexlab#1{#1}\fi

\bibitem[{Tra(2020)}]{Transcript}
 2020.
\newblock \href {https://scrapsfromtheloft.com/stand-up-comedy-scripts/}
  {{Stand-Up Comedy Transcripts}}.

\bibitem[{Alammar(2018)}]{bertembedding}
Jay Alammar. 2018.
\newblock \href {http://jalammar.github.io/illustrated-bert/} {{The Illustrated
  BERT, ELMo, and co. (How NLP Cracked Transfer Learning)}}.

\bibitem[{Annamoradnejad(2020)}]{annamoradnejad2020colbert}
Issa Annamoradnejad. 2020.
\newblock \href {http://arxiv.org/abs/2004.12765} {{ColBERT: Using BERT
  Sentence Embedding for Humor Detection}}.

\bibitem[{Bertero and Fung(2016)}]{bertero-fung-2016-deep}
Dario Bertero and Pascale Fung. 2016.
\newblock \href {https://www.aclweb.org/anthology/L16-1079} {{Deep Learning of
  Audio and Language Features for Humor Prediction}}.
\newblock In \emph{Proceedings of the Tenth International Conference on
  Language Resources and Evaluation ({LREC}'16)}, pages 496--501,
  Portoro{\v{z}}, Slovenia. European Language Resources Association (ELRA).

\bibitem[{Castro et~al.(2018)Castro, Chiruzzo, Ros{\'a}, Garat, and
  Moncecchi}]{castro-etal-2018-crowd}
Santiago Castro, Luis Chiruzzo, Aiala Ros{\'a}, Diego Garat, and Guillermo
  Moncecchi. 2018.
\newblock \href {https://doi.org/10.18653/v1/W18-3502} {{A Crowd-Annotated
  {S}panish Corpus for Humor Analysis}}.
\newblock In \emph{Proceedings of the Sixth International Workshop on Natural
  Language Processing for Social Media}, pages 7--11, Melbourne, Australia.
  Association for Computational Linguistics.

\bibitem[{Cohen(1960)}]{cohen1960}
J.~Cohen. 1960.
\newblock {A Coefficient of Agreement for Nominal Scales}.
\newblock \emph{Educational and Psychological Measurement}, 20(1):37.

\bibitem[{Cohen(1968)}]{cohen1968weighted}
Jacob Cohen. 1968.
\newblock {Weighted kappa: nominal scale agreement provision for scaled
  disagreement or partial credit.}
\newblock \emph{Psychological bulletin}, 70(4):213.

\bibitem[{Devlin et~al.(2018)Devlin, Chang, Lee, and
  Toutanova}]{DBLP:journals/corr/abs-1810-04805}
Jacob Devlin, Ming{-}Wei Chang, Kenton Lee, and Kristina Toutanova. 2018.
\newblock \href {http://arxiv.org/abs/1810.04805} {{BERT:} pre-training of deep
  bidirectional transformers for language understanding}.
\newblock \emph{CoRR}, abs/1810.04805.

\bibitem[{Fleiss et~al.(1971)}]{fleiss1971mns}
J.L. Fleiss et~al. 1971.
\newblock {Measuring nominal scale agreement among many raters}.
\newblock \emph{Psychological Bulletin}, 76(5):378--382.

\bibitem[{Fleiss et~al.(2003)Fleiss, Levin, and Paik}]{articleFleiss}
Joseph Fleiss, Bruce Levin, and Myunghee Paik. 2003.
\newblock \href {https://doi.org/10.1002/0471445428} {{In Statistical Methods
  for Rates and Proportions}}.
\newblock \emph{Statistical Methods for Rates and Proportions}, 203.

\bibitem[{Gillick and Wlodarczak(2019)}]{detection}
Jon Gillick and Marcin Wlodarczak. 2019.
\newblock \href {https://github.com/jrgillick/laughter-detection}
  {laughter-detection}.

\bibitem[{{Graves, Alex and Fernández, Santiago and Schmidhuber,
  Jürgen}(2005)}]{bilstm}
{Graves, Alex and Fernández, Santiago and Schmidhuber, Jürgen}. 2005.
\newblock Bidirectional lstm networks for improved phoneme classification and
  recognition.
\newblock pages 799--804.

\bibitem[{Green(2018)}]{mute}
Jeff Green. 2018.
\newblock \href {https://github.com/jeffgreenca/laughr} {Sitcom laughtrack mute
  tool}.

\bibitem[{Hasan et~al.(2019)Hasan, Rahman, Bagher~Zadeh, Zhong, Tanveer,
  Morency, and Hoque}]{hasan-etal-2019-ur}
Md~Kamrul Hasan, Wasifur Rahman, AmirAli Bagher~Zadeh, Jianyuan Zhong,
  Md~Iftekhar Tanveer, Louis-Philippe Morency, and Mohammed~(Ehsan) Hoque.
  2019.
\newblock \href {https://doi.org/10.18653/v1/D19-1211} {{UR}-{FUNNY}: A
  multimodal language dataset for understanding humor}.
\newblock In \emph{Proceedings of the 2019 Conference on Empirical Methods in
  Natural Language Processing and the 9th International Joint Conference on
  Natural Language Processing (EMNLP-IJCNLP)}, pages 2046--2056, Hong Kong,
  China. Association for Computational Linguistics.

\bibitem[{{Hetzron}(1991)}]{journal1}
R.~{Hetzron}. 1991.
\newblock \href {https://doi.org/https://doi.org/10.1515/humr.1991.4.1.61} {{On
  the structure of punchlines}}.
\newblock \emph{Humor: International Journal of Humor Research}, 4:61--108.

\bibitem[{Krippendorff(2004)}]{krippendorff04}
Klaus Krippendorff. 2004.
\newblock \emph{{Content Analysis: An Introduction to Its Methodology (second
  edition)}}.
\newblock Sage Publications.

\bibitem[{Kuipers(2017)}]{book1}
Giselinde Kuipers. 2017.
\newblock \emph{{In The Anatomy of Laughter}}, chapter Humour styles and class
  cultures: Highbrow humour and lowbrow humour in the netherlands. Routledge.

\bibitem[{Lample and Conneau(2019)}]{lample2019cross}
Guillaume Lample and Alexis Conneau. 2019.
\newblock Cross-lingual language model pretraining.
\newblock \emph{Advances in Neural Information Processing Systems (NeurIPS)}.

\bibitem[{Likert(1932)}]{likert1932technique}
Rensis Likert. 1932.
\newblock {A technique for the measurement of attitudes.}
\newblock \emph{Archives of psychology}.

\bibitem[{Liu et~al.(2019)Liu, Ott, Goyal, Du, Joshi, Chen, Levy, Lewis,
  Zettlemoyer, and Stoyanov}]{DBLP:journals/corr/abs-1907-11692}
Yinhan Liu, Myle Ott, Naman Goyal, Jingfei Du, Mandar Joshi, Danqi Chen, Omer
  Levy, Mike Lewis, Luke Zettlemoyer, and Veselin Stoyanov. 2019.
\newblock \href {http://arxiv.org/abs/1907.11692} {{RoBERTa: {A} Robustly
  Optimized {BERT} Pretraining Approach}}.
\newblock \emph{CoRR}, abs/1907.11692.

\bibitem[{McFee et~al.(2020)McFee, Lostanlen, Metsai, McVicar, Balke, Thomé,
  Raffel, Zalkow, Malek, Dana, Lee, Nieto, Mason, Ellis, Battenberg, Seyfarth,
  Yamamoto, Choi, viktorandreevichmorozov, Moore, Bittner, Hidaka, Wei,
  nullmightybofo, Hereñú, Stöter, Friesch, Weiss, Vollrath, and
  Kim}]{brian_mcfee_2020_3955228}
Brian McFee, Vincent Lostanlen, Alexandros Metsai, Matt McVicar, Stefan Balke,
  Carl Thomé, Colin Raffel, Frank Zalkow, Ayoub Malek, Dana, Kyungyun Lee,
  Oriol Nieto, Jack Mason, Dan Ellis, Eric Battenberg, Scott Seyfarth, Ryuichi
  Yamamoto, Keunwoo Choi, viktorandreevichmorozov, Josh Moore, Rachel Bittner,
  Shunsuke Hidaka, Ziyao Wei, nullmightybofo, Darío Hereñú, Fabian-Robert
  Stöter, Pius Friesch, Adam Weiss, Matt Vollrath, and Taewoon Kim. 2020.
\newblock \href {https://doi.org/10.5281/zenodo.3955228} {librosa/librosa:
  0.8.0}.

\bibitem[{Meaney(2020)}]{meaney-2020-crossing}
J.~A. Meaney. 2020.
\newblock \href {https://doi.org/10.18653/v1/2020.acl-srw.24} {{Crossing the
  Line: Where do Demographic Variables Fit into Humor Detection?}}
\newblock In \emph{Proceedings of the 58th Annual Meeting of the Association
  for Computational Linguistics: Student Research Workshop}, pages 176--181,
  Online. Association for Computational Linguistics.

\bibitem[{Miller et~al.(2020)Miller, Dinh, Simpson, and
  Gurevych}]{miller2020ofaiukp}
Tristan Miller, Erik-Lân~Do Dinh, Edwin Simpson, and Iryna Gurevych. 2020.
\newblock \href {http://arxiv.org/abs/2008.00853} {{OFAI-UKP at
  HAHA@IberLEF2019: Predicting the Humorousness of Tweets Using Gaussian
  Process Preference Learning}}.

\bibitem[{Nolan(2014)}]{film}
Christopher Nolan. 2014.
\newblock Interstellar.

\bibitem[{Park et~al.(2018)Park, Hu, and Muenster}]{Park2018LaughbotDH}
Kate~M. Park, Annie Hu, and Natalie Muenster. 2018.
\newblock {Laughbot: Detecting Humor in Spoken Language with Language and Audio
  Cues}.

\bibitem[{Pennington et~al.(2014)Pennington, Socher, and
  Manning}]{inproceedings}
Jeffrey Pennington, Richard Socher, and Christopher Manning. 2014.
\newblock \href {https://doi.org/10.3115/v1/D14-1162} {{GloVe: Global Vectors
  for Word Representation}}.
\newblock volume~14, pages 1532--1543.

\bibitem[{Rosenthal et~al.(2015)Rosenthal, Bindman, and Randolph}]{book}
A.~Rosenthal, David Bindman, and A.W.B. Randolph. 2015.
\newblock \emph{{No laughing matter: Visual humor in ideas of race,
  nationality, and ethnicity}}.

\bibitem[{Sanh et~al.(2019)Sanh, Debut, Chaumond, and
  Wolf}]{Sanh2019DistilBERTAD}
Victor Sanh, Lysandre Debut, Julien Chaumond, and Thomas Wolf. 2019.
\newblock {DistilBERT, a distilled version of BERT: smaller, faster, cheaper
  and lighter}.
\newblock \emph{ArXiv}, abs/1910.01108.

\bibitem[{Vanbelle(2014)}]{Kappa}
Sophie Vanbelle. 2014.
\newblock \href {https://doi.org/10.1007/s11336-014-9439-4} {{A New
  Interpretation of the Weighted Kappa Coefficients}}.
\newblock \emph{Psychometrika}.

\bibitem[{Weller and Seppi(2019)}]{weller2019humor}
Orion Weller and Kevin Seppi. 2019.
\newblock \href {http://arxiv.org/abs/1909.00252} {{Humor Detection: A
  Transformer Gets the Last Laugh}}.

\bibitem[{Wolf et~al.(2020)Wolf, Debut, Sanh, Chaumond, Delangue, Moi, Cistac,
  Rault, Louf, Funtowicz, Davison, Shleifer, von Platen, Ma, Jernite, Plu, Xu,
  Scao, Gugger, Drame, Lhoest, and Rush}]{wolf-etal-2020-transformers}
Thomas Wolf, Lysandre Debut, Victor Sanh, Julien Chaumond, Clement Delangue,
  Anthony Moi, Pierric Cistac, Tim Rault, Rémi Louf, Morgan Funtowicz, Joe
  Davison, Sam Shleifer, Patrick von Platen, Clara Ma, Yacine Jernite, Julien
  Plu, Canwen Xu, Teven~Le Scao, Sylvain Gugger, Mariama Drame, Quentin Lhoest,
  and Alexander~M. Rush. 2020.
\newblock \href {https://www.aclweb.org/anthology/2020.emnlp-demos.6}
  {{Transformers: State-of-the-Art Natural Language Processing}}.
\newblock In \emph{Proceedings of the 2020 Conference on Empirical Methods in
  Natural Language Processing: System Demonstrations}, pages 38--45, Online.
  Association for Computational Linguistics.

\bibitem[{Yang et~al.(2019{\natexlab{a}})Yang, , {Ai}, and
  Hirschberg}]{Yang20192}
Zixiaofan Yang, , L.~{Ai}, and Julia Hirschberg. 2019{\natexlab{a}}.
\newblock \href {https://doi.org/10.1109/MIPR.2019.00109} {{Multimodal
  Indicators of Humor in Videos}}.
\newblock In \emph{2019 IEEE Conference on Multimedia Information Processing
  and Retrieval (MIPR)}, pages 538--543.

\bibitem[{Yang et~al.(2019{\natexlab{b}})Yang, Hu, and Hirschberg}]{Yang2019}
Zixiaofan Yang, Bingyan Hu, and Julia Hirschberg. 2019{\natexlab{b}}.
\newblock \href {https://doi.org/10.21437/Interspeech.2019-3113} {{Predicting
  Humor by Learning from Time-Aligned Comments}}.
\newblock In \emph{Proc. Interspeech 2019}, pages 496--500.

\end{thebibliography}
\bibliographystyle{acl_natbib}

\end{document}